\pdfoutput=1

\documentclass[11pt]{article}

\usepackage{authblk}
\usepackage{acl}

\usepackage{times}
\usepackage{latexsym}
\usepackage{graphicx}
\usepackage{amsmath}
\usepackage{multirow}

\usepackage[T1]{fontenc}

\usepackage[utf8]{inputenc}
\usepackage{booktabs}

\usepackage{microtype}

\title{Open-Domain Event Graph Induction for Mitigating Framing Bias}

\author[1]{\bf{Siyi Liu}}
\author[2]{\bf{Hongming Zhang}}
\author[2]{\bf{Hongwei Wang}}
\author[2]{\\\bf{Kaiqiang Song}}
\author[1]{\bf{Dan Roth}}
\author[2]{\bf{Dong Yu}}

\affil[1]{University of Pennsylvania}
\affil[2]{Tencent AI Lab, Seattle}

\begin{document}
\maketitle
\begin{abstract}


Researchers have proposed various information extraction (IE) techniques to convert news articles into structured knowledge for news understanding. However, none of the existing methods have explicitly addressed the issue of \textit{framing bias} that is inherent in news articles.
We argue that studying and identifying framing bias is a crucial step towards trustworthy event understanding.
We propose a novel task, \textit{neutral event graph induction}, to address this problem.
An event graph is a network of events and their temporal relations. Our task aims to induce such structural knowledge with minimal framing bias in an open domain. We propose a three-step framework to induce a neutral event graph from multiple input sources. The process starts by inducing an event graph from each input source, then merging them into one merged event graph, and lastly using a Graph Convolutional Network to remove event nodes with biased connotations. We demonstrate the effectiveness of our framework through the use of graph prediction metrics and bias-focused metrics\footnote{Our code and data will be available at https://github.com/liusiyi641/Neutral-Event-Graph}.

\end{abstract}


\begin{figure}
    \centering
    \includegraphics[width=0.5\textwidth]{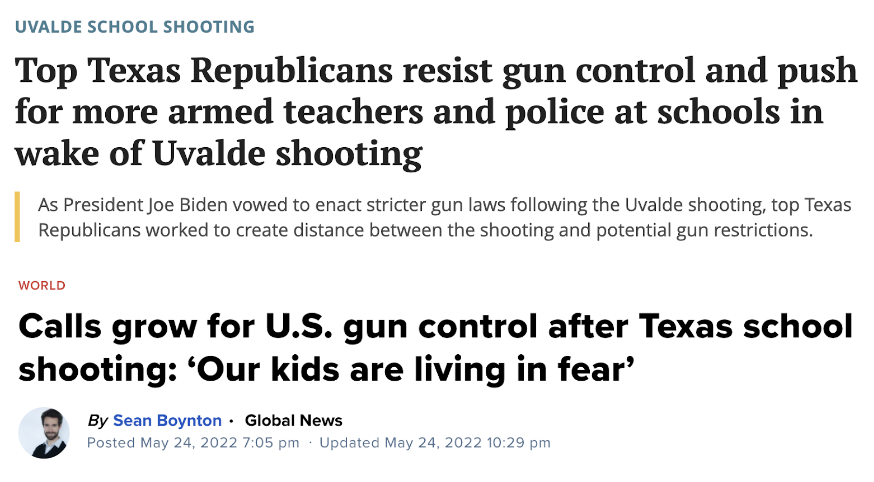}
    \caption{An example of framing bias in news articles. While both news articles are discussing the same event ``Uvalde shooting'', a right-leaning article (top) may argue for increased presence of armed teachers, while a left-leaning article (bottom) may advocate for stricter gun control measures. This illustrates how authors with different political or social beliefs can present the same topic with disparate viewpoints, despite being based on the same set of facts.}
    \label{fig:framing-bias}
\end{figure}

\section{Introduction}

\emph{News editorials} are a form of persuasive text that express the opinions of an editor on a controversial topic. Different authors, with varying political or social beliefs, may present the same topic with distinct events, despite being based on the same set of facts.
For instance, in the case of ``Uvalde shooting'' (Figure \ref{fig:framing-bias}), a right-leaning source (top) may focus on events such as ``Top Republicans resist gun control'', while a left-leaning source (bottom) may highlight events like ``calls for gun control grow''.
This phenomenon is known as \textit{framing bias} \cite{goffman1974frame, Entman, groeling2013media}, where authors can either describe the same event with different linguistic attributes and sentiments (lexical bias), or deliberately include or exclude certain events to promote a particular interpretation (informational bias) \cite{Entman, fan-etal-2019-plain, lee2022neus}.

Researchers have proposed various information extraction (IE) methods for understanding news article, for example, encoding events into \textit{sequences} \cite{schank1977scripts, chambers-jurafsky-2008-unsupervised, chambers-jurafsky-2009-unsupervised, mostafazadeh2016caters, jans2012skip} or \textit{graphs} \cite{wanzare-etal-2016-crowdsourced, li2020connecting, li-etal-2021-future} based on their temporal, causal, and narrative orders.
However, none of these methods have specifically addressed the issue of framing bias or attempted to alleviate this problem.
To address this, we propose a new event representation called \textit{neutral event graph}. A neutral event graph is a network of events (nodes) and their relations (edges) that aim to induce structural knowledge with minimal framing bias. It is noteworthy that our formulation of event graph is distinct from previous graph-based representations \cite{wanzare-etal-2016-crowdsourced, li2020connecting,li-etal-2021-future} in that we do not require a predefined ontology for event types, unlike previous work \cite{li2020connecting} which uses event types to represent nodes. This is because our approach is tailored for the open-domain scenario.

We propose a three-stage approach for neural event graph induction:
(1) \textbf{Event graph induction for single news article}. 
We induce an event graph from each individual news article by using a pretrained salience allocation model to extract the most salient sentences, which are treated as atom events.
The temporal orders among these events are then calculated using an event temporal relation model, resulting in an event graph.
(2) \textbf{Event graphs merging}. We match event nodes across event graphs using a cross-document event co-reference method. This involves merging event graphs on the same side into one graph, followed by further merging these representative graphs into the final neutral graph. A sentence neutralizer is used to rewrite event descriptions when merging coreferential events from different sides, which removes any biased linguistic attributes (i.e. the lexical bias).
(3) \textbf{Framing bias pruning}.
We use graph neural networks to train a binary node classification model that decides which nodes should be removed from the final neutral graph. This helps to alleviate information bias by removing events that are deliberately included to promote certain ideologies.

We have remolded an existing news dataset to facilitate the induction of neutral event graphs. Our comprehensive experiments have demonstrated the effectiveness of our framework through the use of both graph prediction metrics and evaluations that focus on bias.
Our contributions in this paper are as follows:

\begin{itemize}
    \item This is the first study that examines framing bias in event representations and highlights the importance of addressing such bias in event graph induction.
    \item We have proposed a novel method for inducing event graphs, called ``neutral event graph'', which focuses on extracting structured, unbiased knowledge from multiple input documents.
    \item We have developed a three-stage framework and novel evaluation schemes to demonstrate the effectiveness of our approach in reducing framing bias.
\end{itemize}

\begin{figure*}[t]
    \centering
    \includegraphics[width=\textwidth]{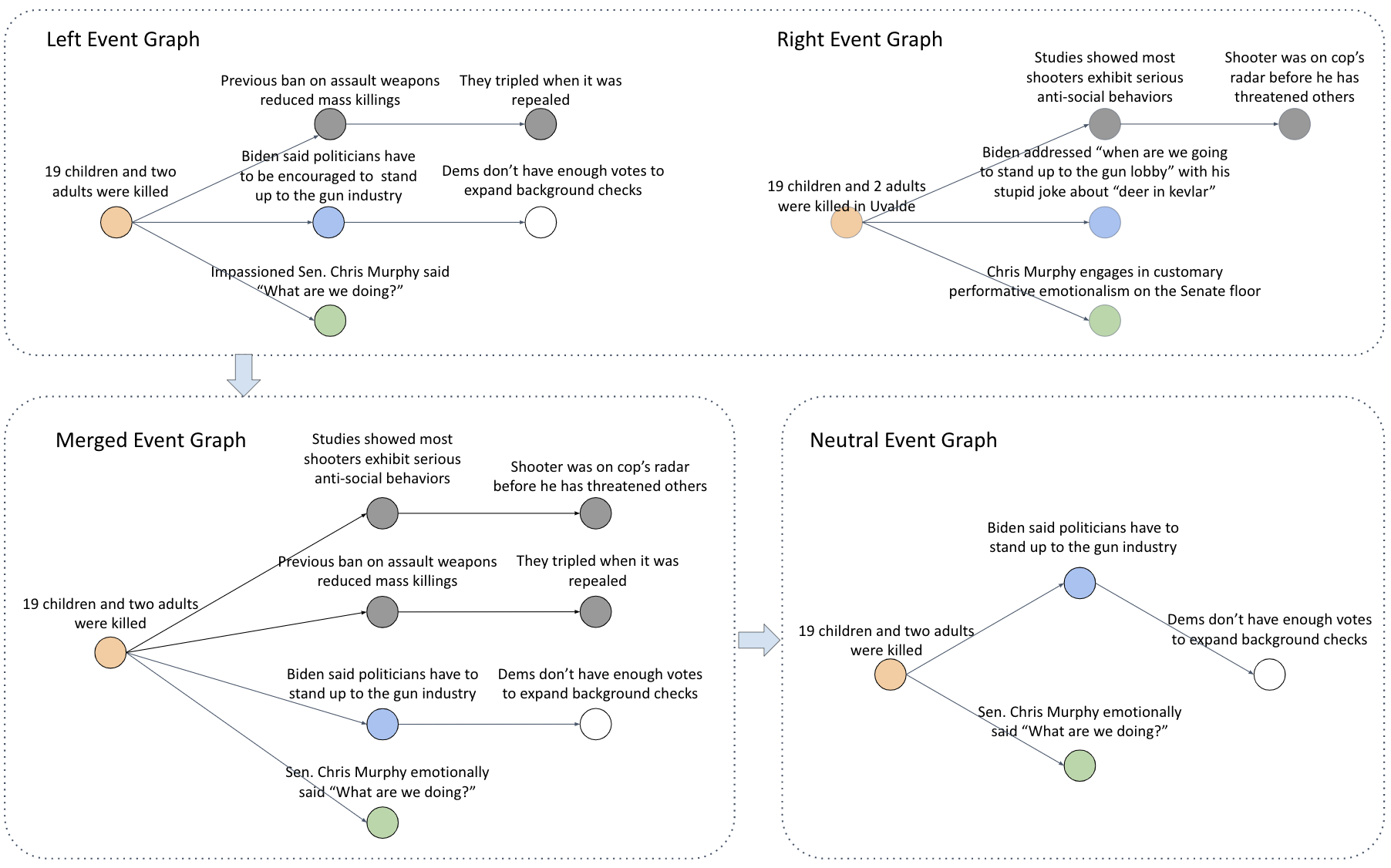}
    \caption{An example of our neutral event graph induction framework. Given two (or more) articles on a topic (Gun control after Uvalde Shooting), we induce an event graph for each article, merge them into a single merged event graph by identifying coreferential event nodes, and use a Graph Convolutional Network to remove biased event nodes and produce a neutral event graph. In this example, the left and right event graphs are induced from a left-leaning and right-leaning news article respectively. The orange, blue, and green nodes represent coreferential nodes and are re-written using our sentence neutralizer to remove lexical bias. The grey nodes represent events that authors deliberately included to sway readers' opinions (informational bias) and are removed by our GCN. The directed edges represent temporal orders.}
    \label{fig:event graph}
\end{figure*}


\section{Related Work}

\subsection{Framing Bias}
Framing is a subtle form of media manipulation in
which some aspects of an issue are highlighted to promote a particular interpretation \cite{goffman1974frame, Entman}.
In a polarized media environment, journalists make
biased decisions regarding which events to cover (informational bias) and how to cover them (lexical bias) to advance certain political agendas \cite{gentzkow2006media, jamieson2007effectiveness, levendusky2013partisan, fan-etal-2019-plain}. In natural language processing (NLP), most previous efforts concerning framing bias focus on automatic bias detection and mitigating it in downstream applications (e.g., summarization) \cite{recasens2013linguistic, yano2010shedding, morstatter2018identifying, liu2019detecting, fan-etal-2019-plain, lee2022neus, hamborg2019illegal}. 

Our proposed task and pipeline focus on mitigating framing bias in event graph induction. Event graph induction methods induce structured knowledge from news articles. However, these news articles can be politically skewed towards certain ideologies or stances, resulting in biased knowledge if not properly mitigated. Our work is the first study that argues for the need to alleviate framing bias in event graph induction and proposes a solution.


\subsection{Event Schema Induction}

Previous efforts in event schema induction focus on set-based, chain-based, and graph-based representations. The set-based methods represent schema as a set where each component is an event trigger \cite{chambers-2013-event, nguyen-etal-2015-generative, huang-ji-2020-semi, yuan, cheung-etal-2013-probabilistic, shen-etal-2021-corpus}. These methods do not model the relations among events. Chain-based representations encode a structured knowledge of prototypical real-life event sequences \cite{schank1977scripts, chambers-jurafsky-2008-unsupervised, chambers-jurafsky-2009-unsupervised, pichotta, rudinger-etal-2015-script}. These representations order the events within a sequence by their event-event relations. Another line of work represents event schemas as graphs \cite{modi-etal-2016-inscript, wanzare-etal-2016-crowdsourced, li-etal-2021-future, li2020connecting, jin-etal-2022-event, weber-etal-2020-causal}. These methods encode global dependencies among atomic events in an entire graph.

Our proposed representation also follows a graph structure. However, we distinguish our graph from the previous graph-based representations in the following aspects: (a). We focus on an open-domain setting, where an ontology for \textit{event types} is undefined. (b). We represent each node as an atomic event itself and connect them with event-event relations, in contrast to \textit{event types} as nodes and entity-entity relations as edges. Our motivation and objective are also different from previous work. We aim at constructing a neutral and global knowledge of a topic by learning from different-sided sources, whereas previous graph-based schemas encode stereotypical structures of events and their connections \cite{li2020connecting, li-etal-2021-future}.



\section{Our Approach}

\subsection{Problem Formulation}
    Suppose that we have a set of news articles $\mathcal A = \{A_1, A_2, \cdots\}$ discussing a specific topic or event.
    These articles are collected from different media sources and may contain incomplete or biased information.
    In this work, we focus on the special case of politics, where a news article can be either left-leaning or right-leaning.
    Therefore, we use $\mathcal A = \{A_{l_1}, A_{l_2}, \cdots, A_{r_1}, A_{r_2}, \cdots\}$ to represent the set of news articles, where $A_{l_i}$/$A_{r_i}$ is the $i$-th article on the left/right side.
    Our goal is to construct a \textit{neutral event graph}, $G_{neutral}$, which covers all the information conveyed in $\mathcal A$, while eliminating any framing bias, including lexical bias and informational bias.
    In $G_{neutral}$, nodes represent atom events and edges represent temporal relations between them.
    
    As we are working in the open-domain scenario, we do not predefine an ontology for the nodes, but instead represent them using sentences or phrases, which provides richer linguistic features for describing complex events.

\subsection{Single Document Event Graph Induction}
\label{sec:induction}
The first step of our proposed method is to construct an event graph for a news article.
To do this, we first extract salient events from the input article.
Traditional event extraction methods rely on human-labeled event types as the supervision for their models.
However, in the open-domain setting, we don't have any predefined ontology for event types.
To overcome this challenge, we use a salience prediction model to determine the salience of events.
Specifically, given an input news article, we first use \textsc{Season} \cite{feiwang}, a transformer-based abstractive summarization model for identifying the most salient sentences in the new article.
Then select the top-$k$ sentences with the highest salience scores are served as the atom events.

Inspired by \cite{shen-etal-2021-corpus}, a dependency parser is used to extract all subject-verb-object (SVO) triplets for the selected sentences.
In addition, we use an off-the-shelf temporal relation prediction tool \cite{wang2020joint} to predict the temporal relation between atom events.
It takes the extracted SVO triplets as input and predicts the temporal relations between events as directed edges in the event graph.
Finally, we convert the graph into a directed acyclic graph (DAG) by repeatedly removing the edge with the lowest confidence score until there is no cycle left in the graph.

%

\subsection{Event Graph Merging}
\label{sec:merging}
We have constructed the set of event graphs $\mathcal G=\{G_{l_1}, G_{l_2}, \cdots, G_{r_1}, G_{r_2}, \cdots\}$ from input articles $\mathcal A = \{A_{l_1}, A_{l_2}, \cdots, A_{r_1}, A_{r_2}, \cdots\}$.
The next step is to merge these event graphs together.

We first merge the graphs on the same side into two intermediate graphs: $G_{left}$ and $G_{right}$.
To accomplish this, we use an event coreference detection tool \cite{yu2020paired} to match the event nodes in two graphs.
Specifically, to merge two graphs $G_{l_i}$ and $G_{l_j}$, we calculate the matching score of each node $v \in G_{l_i}$ with all nodes in $G_{l_j}$, and select the node with the highest score as the coreferential node for $v$ if the score exceeds a predefined threshold.
We then randomly choose one of the coreferential nodes to represent the merged node.
In this way, $G_{l_i}$ and $G_{l_j}$ can be merged into a single graph, which is then converted into a DAG using the same post-processing method described in Section \ref{sec:induction}.

The above step is repeated until we obtain a single graph $G_{left}$.
Similarly,  $G_{right}$ can also be obtained using the same process.
The final step is to merge $G_{left}$ and $G_{right}$ into $G$.
This process is similar to the merging procedure described earlier, with the exception of using a sentence-level neutralizer when merging two coreferential nodes.
Specifically, the neutralizer takes the coreferential nodes as input and generate a less biased sentence as the merged node.
This is necessary because nodes from different sides do not typically share similar linguistic attributes, even if they are coreferential.
Therefore, a pretrained neutralizer is used to rewrite their content into a neural sentence.


 \begin{figure*}
    \centering
    \includegraphics[width=0.9\textwidth]{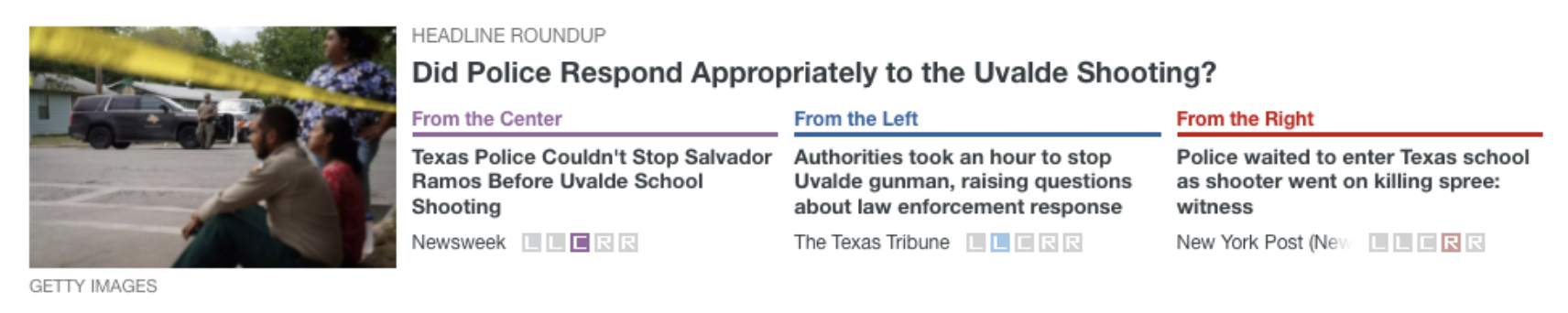}
    \caption{A example triplet from \textsc{Allsides.com}}
    \label{fig:triplet}
\end{figure*}

\subsection{Framing Bias Pruning}
The graph $G$ is obtained by merging $G_{left}$ and $G_{right}$ together.
However, the resulting merged graph is often larger than a typical event graph and contains a significant proportion of nodes that are not relevant to the main aspect of the complex event.
Thus, it is necessary to prone the merged graph and remove these unimportant nodes in order to effectively analyze the key aspect of the complext event.

One potential solution to this issue is to train a binary node classification model on the merged graph $G$, which can distinguish between important and unimportant nodes.
However, there is a lack of ground truth node labels to train this model.
As a solution, we realize that in addition to collecting left-leaning and right-leaning articles on the same topic, we can also gather central-leaning articles (more information can be found in Section \ref{sec:dataset}).
These central-leaning articles can serve as pseudo ground truth for the pruning procedure.

Specifically, we process central-leaning news articles and construct a graph, $G_{central}$, using the same method described in Section \ref{sec:induction} and \ref{sec:merging}.
We then calculate the exact node matching between $G$ and $G_{central}$ by maximizing the following objective:
\begin{equation}
    \begin{split}
        &\max \sum_{v_i \in G, v_j \in G_{central}} sim(v_i, v_j) \cdot A_{ij} \\
        s.t. \ & A_{ij} \in \{0, 1\}, \ \ \text{for all} \ i \ \text{and} \ j, \\
        & \sum_i A_{ij} \leq 1, \ \ \text{for all} \ j, \\
        & \sum_j A_{ij} \leq 1, \ \ \text{for all} \ i,
    \end{split}
\end{equation}
where $sim(v_i, v_j)$ is the cosine similarity between the embeddings of $v_i$ and $v_j$'s content.
We solve this problem by greedily selecting a pair of nodes $(v_i, v_j)$ with the highest similarity score from the two graphs, as long as they have not been matched to any node yet.
Then the nodes $v_i$ and $v_j$ are matched if their similarity score exceeds a pre-defined threshold.
This procedure is repeated until either one of the graphs is fully matched or the remaining similarity scores fall below the threshold.
Through this method, the matched nodes in $G$ can be considered as important, while the unmatched nodes can be viewed as unimportant.
These labels can then be used to train the binary node classification model.

We use Graph Convolutional Network (GCN) as the implementation for the binary node classification model.
GCN learns the representation of each node by aggregating the representations of its' neighbors \cite{kipf2016semi}.
Here we employ a conventional 2-layer GCN structure:
\begin{equation}
    Y = \mathrm{softmax}(\hat{A}\, \mathrm{ReLU}(\hat{A}XW^{(0)})W^{(1)}),
\end{equation}
where $X$ is the initial node embeddings of $G$, $\hat{A}$ is the normalized adjacency matrix of $G$, $W^{(0)}$ and $W^{(1)}$ are the weight matrices of GCN, and $Y$ is the predicted labels indicating whether a node should be kept in the graph or not.

\section{Experiments}
\subsection{Dataset}
\label{sec:dataset}
Currently, there are no existing news  datasets that provide supervision for sources from different sides of a topic. To address this, we have created our own dataset by extending a multi-document summarization corpus. The dataset, called \textsc{NeuS}, contains 3,564 triplets of news articles from \textsc{Allsides.com} \cite{lee2022neus}. Each triplet includes three news articles from left, right, and center-leaning American publishers on the same event. The dataset is in English and primarily focuses on U.S. political topics. An example triplet is illustrated in Figure \ref{fig:triplet}.

We extract the text contents of each news article of every triplet from \textsc{NeuS} using the article links provided by \citet{lee2022neus}.\footnote{We use NewsPaper3k library (\url{https://newspaper.readthedocs.io/en/latest/}) to extract the text contents.} Due to stale and broken links, this results in 1,766 valid triplets of news articles. For each triplet of news, we induce an event graph from the center news article following the same protocol as in Section \ref{sec:induction}. We can then consider this center event graph as our target graph and train a system to induce it from a pair of left and right-leaning news articles on the same issue. It's worth noting that the term ``center'' in this context does not imply completely framing bias-free. ``Center'' news outlets are usually less tied to a particular political ideology, which means they are less likely to frame the article in a particular way to promote certain political interpretations. However, their reports may still contain framing bias because editorial judgement naturally leads to human-induced biases \cite{lee2022neus}.


\subsection{Baselines} 
\paragraph{Left and Right Event Graphs.}
The \textsc{Left} and \textsc{Right} baselines refer to the event graphs, $G_{left}$ and $G_{right}$, which are induced from the left-leaning and right-leaning articles, respectively, following the induction and merging process outlined in Section \ref{sec:induction} and \ref{sec:merging}.

\paragraph{Salience Ranking Model.} Our first baseline is a salience-based event induction model. Specifically, we concatenate the input articles into one, and extract events based on a salience metric. We then induce an event graph using the same temporal prediction tool. We adopt the method used in \cite{shen-etal-2021-corpus} and compute the salience score of a word (a predicate lemma or an object head) as follows:
\begin{align*}
    \textit{Salience(w)} &= (1+\log freq(w))^2 \log\frac{N}{bsf(w)},
\end{align*}
where $freq(w)$ is the frequency of the word $w$, $N$ is the number of sentences in a background corpus, and $bsf(w)$ is the background sentence frequency of the word $w$. The concept is similar to TF-IDF and we use the English Wikipedia 20171201 dump as our background corpus, as done in \cite{shen-etal-2021-corpus}.

\paragraph{Event Instance Graph Model.} \citet{li-etal-2021-future} propose an auto-regressive graph generation model that learns to generate the next event type node with its argument. However, in our setting, we don't have any information on the event type and entity level, so we only adapt their procedure of constructing \textit{event instance graphs} to our setting. Specifically, we extract the events and their temporal relations for each input article on a topic, and construct one \textit{event instance graph} for the topic by merging all coreferential events nodes. \citet{li-etal-2021-future} consider the isolated nodes as irrelevant and exclude them in the instance graph, whereas we experiment with both including and excluding these isolated nodes in our study.

\subsection{Experimental Details}
We divide the dataset into train/val/test splits with 70\% of the data being used for training, 10\% for validation, and 20\% for testing. This results in 1,236 instances for training, 176 instances for validation, and 354 instances for testing.


\paragraph{Dependency Parser.}
We follow the method used in \cite{shen-etal-2021-corpus} and use the Spacy \textit{$en\_core\_web\_lg$} tool as our dependency parser for extracting subject-verb-objects (SVOs).

\paragraph{Salience Allocation.}
We use the model \textsc{Season} to predict the salience score for each sentence in an article \cite{feiwang}. \textsc{Season} is a transformer-based abstractive summarization model that incorporates salience prediction and text summarization into a single network. During training, the model jointly learns to predict the degree of salience for each sentence and is guided by ROUGE-based ground-truth salience allocation to generate the abstractive summary. \textsc{Season} was trained on the CNN/DM dataset \cite{CNNDM} and achieved 43.08 RougeL performance within the domain.

\paragraph{Event Temporal Relation.} 
\citet{wang2020joint} proposed a joint-constrained learning framework for predicting event-event relations. We adopt the same framework and use a model trained on the MATRES dataset \cite{ning-etal-2018-improving} as our event temporal relation prediction model. The model achieved 78.8 F1 in the MATRES dataset.

\paragraph{Event Coreference.}
\textsc{PairwiseRL} is a pair-wise representation learning scheme for event mention pairs \cite{yu2020paired}. We use a model trained on ECB+ \cite{cybulska-vossen-2014-using}, a cross-document event coreference dataset, to identify coreferential events nodes across documents.

\paragraph{Sentence Neutralizer.}
We fine-tune a pretrained BART model \cite{lewis2019bart} using the titles in the \textsc{NeuS} dataset as our sentence neutralizer. The goal of the model is to generate a neutral sentence (the center article's title) given the left and right articles' titles. We chose to use the titles of the news articles as our training data because (1) the title is in the same domain, (2) the title is roughly the same length as an event sentence, and (3) using the titles will not contaminate our evaluation as the titles are excluded from the articles' contents. We trained a bart-large model with 12 encoder and decoder layers for 6 epochs with a learning rate of 1e-7. It achieved 32.96 RougeL score on validation (10\%).

\paragraph{Graph Convolutional Network.}
\label{sec:gnn}
We train a 2-layer Graph Convolutional Network (GCN) following the method of \citet{kipf2016semi} as our node classification model to decide whether to remove an event node. The node feature representations for each event are initialized using SimCSE sentence embeddings with a dimension of 768 \cite{gao2021simcse}. We train it for 10 epochs with a learning rate of 1e-4.

\begin{table*}[t]
\begin{tabular} {c|ccc|ccc}
\toprule
 \multirow{2}{*}{Model} & \multicolumn{3}{c|}{Node-level} & \multicolumn{3}{c}{Edge-level} \\ 

 & \textsc{Precision} & \textsc{Recall} & \textsc{F1} & \textsc{Precision} & \textsc{Recall} & \textsc{F1} \\
\midrule
\textsc{Left} & 49.31 & 53.64 & 48.57 & 35.87 & 40.10 & 31.72 \\
\textsc{Right} & 54.38& 51.53& 50.35 & 39.54& 35.23 & 32.53 \\ \hline
Salience Ranking &31.26 & 37.98 & 32.09 & 20.11 & 31.13 & 19.37 \\ 
Event Instance Graph & 62.78 & 28.40 & 33.22 & 34.63 & 30.22 & 27.56 \\
w/ isolated nodes  & 38.98 & 54.88 & 42.13 & 30.01 &49.19  & 31.69 \\ 
\hline
\textsc{Ours}  & \textbf{64.30} & \textbf{57.97} & \textbf{59.08} & \textbf{51.31} & \textbf{52.79}& \textbf{48.47} \\
\bottomrule
\end{tabular}
\centering
\caption{Evaluation results with graph distance metrics.}
\label{tab:result}
\end{table*}

\subsection{Evaluation Metrics}
We propose three metrics to evaluate our framework. The first two metrics evaluate the generated neutral event graph by computing the distance between it and the target center event graph. The last metric evaluates the degree of bias of the neutral event graph in relation to the target center graph.

\subsubsection{Graph Distance Metrics}
We evaluate the quality of the generated event graphs by measuring their distance to the center event graphs, which are considered as the target graphs with very little framing bias given their center-leaning ideologies.
We propose two distance metrics to compare the generated event graph with the target center graph. One key challenge here is to define a distance metric between two event nodes. Most previous studies determine if two events match by simply comparing whether they have the same event types \cite{li2020connecting, li-etal-2021-future, jin-etal-2022-event}. However, in our open-domain setting, we don't have a predefined ontology for event types, so we make use of pre-trained sentence embeddings as our method to compare events. Specifically, for a pair of event sentences, we embed them using SimCSE \cite{gao2021simcse} and decide that these two events match if the cosine similarity between their sentence embeddings is above 0.5.

\paragraph{Node-Level Metric.}
We compute the pair-wise cosine similarities between all the event nodes in our generated neutral graph and the target center graph. We then greedily take a pair of event nodes with the highest similarity score, count it as a true positive, and remove them from both graphs. We iterate this process until there is no pair of events that have a similarity score higher than 0.5 or we run out of nodes in either graph. We then consider the rest of the nodes left in the predicted neutral graph and target center graph as false positives and false negatives, respectively. Finally, we compute precision, recall, and F1 scores based on these numbers as our final node-level metrics.

\paragraph{Edge-Level Metric.}
We propose a stricter metric that evaluates the similarity of edges between two graphs. Similar to the node-level metric, we match edges instead of nodes. For each edge $(u_p,v_p)$ in the predicted graph and each edge $(u_t, v_t)$ in the target graph, we calculate the similarity as $Similarity= (Sim(u_p,u_t) + Sim(u_p, v_t))/2$. We consider an edge to be a true positive if its similarity is above 0.5, and iterate until no valid matches can be found. The edge-level precision, recall, and F1 are then calculated in the same way as the node-level metric.

\subsubsection{Bias Metric}
We suggest a lexicon-based polarity metric to evaluate the degree of framing bias (more specifically, lexical bias) of our generated neutral event graph, following a similar procedure in \cite{lee2022neus}. 

Valence-Arousal-Dominance (VAD) \cite{mohammad-2018-obtaining} dataset has a large list of lexicons annotated with their valence (v), arousal (a) and dominance (d) scores. Valence, arousal, and dominance represent the direction of polarity (positive, negative), the strength of the polarity (active, passive), and the level of control (powerful, weak), respectively. We use the valence score to determine whether a token is biased, and the arousal score to determine how biased the token is.

Specifically, for each event $u_i = [t_1, t_2, ..., t_n]$, $u_i \in V_{neutral}$ from our generated neutral event graph $G_{neutral}$, we extract every token $t_k$ and add it to a set $S_{neutral}$. We do the same for each event $v_j \in V_{center}$ for the center graph $G_{center}$ and get a set of tokens $S_{center}$. We then filter out all tokens in $S_{neutral}$ that are present in $S_{center}$, such that 
\[S^*_{neutral} \subseteq S_{neutral}\] and 
\[
S^*_{neutral} \cap S_{center} = \emptyset. \]
This ensures that we are measuring
the relative polarity of $V_{neutral}$ in reference to the target $V_{center}$. Then for each token in $S^*_{neutral}$, we select tokens with either positive valence (v > 0.65) or negative valence (v < 0.35) to eliminate neutral words (e.g., stopwords and non-emotion-provoking words). We then sum the arousal scores for the identified positive and negative tokens and get Arousal$_{pos}$ and Arousal$_{neg}$ for the set $S^*_{neutral}$. We average these arousal scores across the test-set as our final metric.

This metric approximates the existence of lexical bias by quantifying how arousing and sensational the events are in our generated neutral event graph.


\subsection{Results and Discussion}
Table \ref{tab:result} demonstrates our results evaluated using graph distance metrics. We see that the neutral event graph generated using our framework improves from the baselines by large margins. We also notice that Event Instance Graph model generates a graph with high node-level precision but low recall. This is because it only keeps the coreferential nodes and excludes other isolated nodes. The coreferential nodes are more likely to be present in $G_{center}$. But excluding all isolated nodes will fail to preserve some isolated nodes that are also salient to $G_{center}$, for instance the white node in Figure \ref{fig:event graph}, and therefore leads to a low recall. Similar intuitions can be found with the Event Instance Graph that includes all isolated nodes. It leads to a high recall but low precision. Our framework, on the other hand, alleviates this problem by learning a GCN to predict whether a node should be removed or not. This flexibility provides us gain in both precision and recall. Besides, our framework uses a GCN to learn the structure of the graphs, which offers us substantial gains in edge-level metrics. 

We present our results with the bias-level metric in table \ref{tab:polarity}. We observe that using our framework, we can mitigate the lexical bias of our generated event graphs from the original left-leaning and right-leaning event graphs. It shows that the event graph induced by our framework contains less events with sensational attributes that are used to promote certain perceptions. Besides, we show that our sentence neutralizer can help remove the arousing linguistic features from their input events by conducting an ablation experiment that excludes the neutralizer in our merging step. 

We also experiment with different neural network structures in our pruning stage to examine the effect of GCN in our framework. We discover that training a multi-layer perceptron (MLP) on the node features (without any graph-level information) can do fairly well at the task evaluated by both node-level and edge-level metrics. Graph Attention Networks (GAT), on the other hand, performs well under the node F1 with an attention mechanism on the node features, but fails to maintain a good graph structure as shown by the lower edge F1 score. The GCN model exceeds both the other two models by learning a convolution over neighbourhoods.
All of the models here are trained under the same settings as described in \ref{sec:gnn} with the same random seed.

\begin{table}[t]
    \centering
    \begin{tabular}{lcc}
        \toprule
         Graph & Arousal$_{pos}$ & Arousal$_{neg}$  
         \\
         \midrule
         \textsc{Left} & 10.97 & 6.31  \\
         \textsc{Right} & 8.96 & 5.20 \\
         \hline
         \textsc{Ours} & \textbf{6.12} & \textbf{3.60}  \\
         \,\, w/o Neutralizer &7.75 &4.49 \\
         \bottomrule
    \end{tabular}
    \caption{Evaluation results with bias metric. The lower the arousal score, the less biased a graph is. w/o Neutralizer means to replace the sentence neutralizer in our framework with a random choice from the input events.}
    \label{tab:polarity}
\end{table}

\begin{table}[t]
    \centering
    \begin{tabular}{lcc}
        \toprule
         Pruning Model & Node F1 & Edge F1  
         \\
         \midrule
         \textsc{MLP} & 56.72 & 44.67  \\

         \textsc{GAT} & 58.44 & 42.90 \\
         \hline
         \textsc{GCN} & \textbf{59.08} & \textbf{48.47} \\
         \bottomrule
    \end{tabular}
    \caption{Graph distance metric results of using different neural networks as our pruning model.}
    \label{tab:ablation}
\end{table}

\section{Conclusion}
In this work, we propose a novel task called "neutral event graph induction" which aims to create a "neutral event graph" that has minimal framing bias from multiple input articles. We present a three-step process to achieve this. The process starts by inducing event graphs from each input document, then merging them based on their stances and finally pruning the merged graph by eliminating biased event nodes. Our experiments demonstrate the effectiveness of our framework using both graph distance metrics and framing bias metrics. In the future, we plan to expand our framework and experiments to a more diverse setting, covering different topics, stances, and domains.

\section{Limitations}
One limitation of our task's setup is that there may be events that are significant to center-leaning articles, but are not covered or discussed in both left and right-leaning articles. This creates an information bottleneck for the task, which could be potentially addressed by including more left and right-leaning input articles.
Another limitation is that the central-leaning articles do not necessarily imply a complete lack of framing bias. ``Central'' news outlets are usually less closely associated with a particular political ideology, but their reports may still contain framing bias due to the inherent biases that result from editorial judgment.

\section{Ethical Considerations}
The articles in our dataset are written by professional journalists for the purpose of promoting political interpretations. However, in rare cases, the authors may use extremely sensational language or inappropriate and aggressive language for the purpose of political propaganda.

\section{Impact}
Framing bias is inherent in news articles, particularly in editorials on controversial topics. Our aim is to bring attention to the recognition and mitigation of framing bias not only in news articles but also in downstream applications such as event graph induction. We hope that our research can serve as a valuable resource for other researchers in this field.






\bibliography{anthology,custom}

\begin{thebibliography}{42}
\expandafter\ifx\csname natexlab\endcsname\relax\def\natexlab#1{#1}\fi

\bibitem[{Chambers(2013)}]{chambers-2013-event}
Nathanael Chambers. 2013.
\newblock \href {https://aclanthology.org/D13-1185} {Event schema induction
  with a probabilistic entity-driven model}.
\newblock In \emph{Proceedings of the 2013 Conference on Empirical Methods in
  Natural Language Processing}, pages 1797--1807, Seattle, Washington, USA.
  Association for Computational Linguistics.

\bibitem[{Chambers and Jurafsky(2008)}]{chambers-jurafsky-2008-unsupervised}
Nathanael Chambers and Dan Jurafsky. 2008.
\newblock \href {https://aclanthology.org/P08-1090} {Unsupervised learning of
  narrative event chains}.
\newblock In \emph{Proceedings of ACL-08: HLT}, pages 789--797, Columbus, Ohio.
  Association for Computational Linguistics.

\bibitem[{Chambers and Jurafsky(2009)}]{chambers-jurafsky-2009-unsupervised}
Nathanael Chambers and Dan Jurafsky. 2009.
\newblock \href {https://aclanthology.org/P09-1068} {Unsupervised learning of
  narrative schemas and their participants}.
\newblock In \emph{Proceedings of the Joint Conference of the 47th Annual
  Meeting of the {ACL} and the 4th International Joint Conference on Natural
  Language Processing of the {AFNLP}}, pages 602--610, Suntec, Singapore.
  Association for Computational Linguistics.

\bibitem[{Cheung et~al.(2013)Cheung, Poon, and
  Vanderwende}]{cheung-etal-2013-probabilistic}
Jackie Chi~Kit Cheung, Hoifung Poon, and Lucy Vanderwende. 2013.
\newblock \href {https://aclanthology.org/N13-1104} {Probabilistic frame
  induction}.
\newblock In \emph{Proceedings of the 2013 Conference of the North {A}merican
  Chapter of the Association for Computational Linguistics: Human Language
  Technologies}, pages 837--846, Atlanta, Georgia. Association for
  Computational Linguistics.

\bibitem[{Cybulska and Vossen(2014)}]{cybulska-vossen-2014-using}
Agata Cybulska and Piek Vossen. 2014.
\newblock \href
  {http://www.lrec-conf.org/proceedings/lrec2014/pdf/840_Paper.pdf} {Using a
  sledgehammer to crack a nut? lexical diversity and event coreference
  resolution}.
\newblock In \emph{Proceedings of the Ninth International Conference on
  Language Resources and Evaluation ({LREC}'14)}, pages 4545--4552, Reykjavik,
  Iceland. European Language Resources Association (ELRA).

\bibitem[{Entman(1993)}]{Entman}
Robert~M. Entman. 1993.
\newblock \href {https://doi.org/10.1111/j.1460-2466.1993.tb01304.x} {{Framing:
  Toward Clarification of a Fractured Paradigm}}.
\newblock \emph{Journal of Communication}, 43(4):51--58.

\bibitem[{Fan et~al.(2019)Fan, White, Sharma, Su, Choubey, Huang, and
  Wang}]{fan-etal-2019-plain}
Lisa Fan, Marshall White, Eva Sharma, Ruisi Su, Prafulla~Kumar Choubey, Ruihong
  Huang, and Lu~Wang. 2019.
\newblock \href {https://doi.org/10.18653/v1/D19-1664} {In plain sight: Media
  bias through the lens of factual reporting}.
\newblock In \emph{Proceedings of the 2019 Conference on Empirical Methods in
  Natural Language Processing and the 9th International Joint Conference on
  Natural Language Processing (EMNLP-IJCNLP)}, pages 6343--6349, Hong Kong,
  China. Association for Computational Linguistics.

\bibitem[{Gao et~al.(2021)Gao, Yao, and Chen}]{gao2021simcse}
Tianyu Gao, Xingcheng Yao, and Danqi Chen. 2021.
\newblock Simcse: Simple contrastive learning of sentence embeddings.
\newblock \emph{arXiv preprint arXiv:2104.08821}.

\bibitem[{Gentzkow and Shapiro(2006)}]{gentzkow2006media}
Matthew Gentzkow and Jesse~M Shapiro. 2006.
\newblock Media bias and reputation.
\newblock \emph{Journal of political Economy}, 114(2):280--316.

\bibitem[{Goffman(1974)}]{goffman1974frame}
Erving Goffman. 1974.
\newblock \emph{Frame analysis: An essay on the organization of experience.}
\newblock Harvard University Press.

\bibitem[{Groeling(2013)}]{groeling2013media}
Tim Groeling. 2013.
\newblock Media bias by the numbers: Challenges and opportunities in the
  empirical study of partisan news.
\newblock \emph{Annual Review of Political Science}, 16(1):129--151.

\bibitem[{Hamborg et~al.(2019)Hamborg, Zhukova, and Gipp}]{hamborg2019illegal}
Felix Hamborg, Anastasia Zhukova, and Bela Gipp. 2019.
\newblock Illegal aliens or undocumented immigrants? towards the automated
  identification of bias by word choice and labeling.
\newblock In \emph{International Conference on Information}, pages 179--187.
  Springer.

\bibitem[{Huang and Ji(2020)}]{huang-ji-2020-semi}
Lifu Huang and Heng Ji. 2020.
\newblock \href {https://doi.org/10.18653/v1/2020.emnlp-main.53}
  {Semi-supervised new event type induction and event detection}.
\newblock In \emph{Proceedings of the 2020 Conference on Empirical Methods in
  Natural Language Processing (EMNLP)}, pages 718--724, Online. Association for
  Computational Linguistics.

\bibitem[{Jamieson et~al.(2007)Jamieson, Hardy, and
  Romer}]{jamieson2007effectiveness}
Kathleen~Hall Jamieson, Bruce~W Hardy, and Daniel Romer. 2007.
\newblock The effectiveness of the press in serving the needs of american
  democracy.

\bibitem[{Jans et~al.(2012)Jans, Bethard, Vulic, and Moens}]{jans2012skip}
Bram Jans, Steven Bethard, Ivan Vulic, and Marie-Francine Moens. 2012.
\newblock Skip n-grams and ranking functions for predicting script events.
\newblock In \emph{Proceedings of the 13th Conference of the European Chapter
  of the Association for Computational Linguistics (EACL 2012)}, pages
  336--344. ACL; East Stroudsburg, PA.

\bibitem[{Jin et~al.(2022)Jin, Li, and Ji}]{jin-etal-2022-event}
Xiaomeng Jin, Manling Li, and Heng Ji. 2022.
\newblock \href {https://doi.org/10.18653/v1/2022.naacl-main.147} {Event schema
  induction with double graph autoencoders}.
\newblock In \emph{Proceedings of the 2022 Conference of the North American
  Chapter of the Association for Computational Linguistics: Human Language
  Technologies}, pages 2013--2025, Seattle, United States. Association for
  Computational Linguistics.

\bibitem[{Kipf and Welling(2016)}]{kipf2016semi}
Thomas~N Kipf and Max Welling. 2016.
\newblock Semi-supervised classification with graph convolutional networks.
\newblock \emph{arXiv preprint arXiv:1609.02907}.

\bibitem[{Lee et~al.(2022)Lee, Bang, Yu, Madotto, and Fung}]{lee2022neus}
Nayeon Lee, Yejin Bang, Tiezheng Yu, Andrea Madotto, and Pascale Fung. 2022.
\newblock Neus: Neutral multi-news summarization for mitigating framing bias.

\bibitem[{Levendusky(2013)}]{levendusky2013partisan}
Matthew~S Levendusky. 2013.
\newblock Why do partisan media polarize viewers?
\newblock \emph{American Journal of Political Science}, 57(3):611--623.

\bibitem[{Lewis et~al.(2019)Lewis, Liu, Goyal, Ghazvininejad, Mohamed, Levy,
  Stoyanov, and Zettlemoyer}]{lewis2019bart}
Mike Lewis, Yinhan Liu, Naman Goyal, Marjan Ghazvininejad, Abdelrahman Mohamed,
  Omer Levy, Ves Stoyanov, and Luke Zettlemoyer. 2019.
\newblock Bart: Denoising sequence-to-sequence pre-training for natural
  language generation, translation, and comprehension.
\newblock \emph{arXiv preprint arXiv:1910.13461}.

\bibitem[{Li et~al.(2021)Li, Li, Wang, Huang, Cho, Ji, Han, and
  Voss}]{li-etal-2021-future}
Manling Li, Sha Li, Zhenhailong Wang, Lifu Huang, Kyunghyun Cho, Heng Ji,
  Jiawei Han, and Clare Voss. 2021.
\newblock \href {https://doi.org/10.18653/v1/2021.emnlp-main.422} {The future
  is not one-dimensional: Complex event schema induction by graph modeling for
  event prediction}.
\newblock In \emph{Proceedings of the 2021 Conference on Empirical Methods in
  Natural Language Processing}, pages 5203--5215, Online and Punta Cana,
  Dominican Republic. Association for Computational Linguistics.

\bibitem[{Li et~al.(2020)Li, Zeng, Lin, Cho, Ji, May, Chambers, and
  Voss}]{li2020connecting}
Manling Li, Qi~Zeng, Ying Lin, Kyunghyun Cho, Heng Ji, Jonathan May, Nathanael
  Chambers, and Clare Voss. 2020.
\newblock Connecting the dots: Event graph schema induction with path language
  modeling.
\newblock In \emph{Proceedings of the 2020 Conference on Empirical Methods in
  Natural Language Processing (EMNLP)}, pages 684--695.

\bibitem[{Liu et~al.(2019)Liu, Guo, Mays, Betke, and Wijaya}]{liu2019detecting}
Siyi Liu, Lei Guo, Kate Mays, Margrit Betke, and Derry~Tanti Wijaya. 2019.
\newblock Detecting frames in news headlines and its application to analyzing
  news framing trends surrounding us gun violence.
\newblock In \emph{Proceedings of the 23rd conference on computational natural
  language learning (CoNLL)}.

\bibitem[{Modi et~al.(2016)Modi, Anikina, Ostermann, and
  Pinkal}]{modi-etal-2016-inscript}
Ashutosh Modi, Tatjana Anikina, Simon Ostermann, and Manfred Pinkal. 2016.
\newblock \href {https://aclanthology.org/L16-1555} {{I}n{S}cript: Narrative
  texts annotated with script information}.
\newblock In \emph{Proceedings of the Tenth International Conference on
  Language Resources and Evaluation ({LREC}'16)}, pages 3485--3493,
  Portoro{\v{z}}, Slovenia. European Language Resources Association (ELRA).

\bibitem[{Mohammad(2018)}]{mohammad-2018-obtaining}
Saif Mohammad. 2018.
\newblock \href {https://doi.org/10.18653/v1/P18-1017} {Obtaining reliable
  human ratings of valence, arousal, and dominance for 20,000 {E}nglish words}.
\newblock In \emph{Proceedings of the 56th Annual Meeting of the Association
  for Computational Linguistics (Volume 1: Long Papers)}, pages 174--184,
  Melbourne, Australia. Association for Computational Linguistics.

\bibitem[{Morstatter et~al.(2018)Morstatter, Wu, Yavanoglu, Corman, and
  Liu}]{morstatter2018identifying}
Fred Morstatter, Liang Wu, Uraz Yavanoglu, Stephen~R Corman, and Huan Liu.
  2018.
\newblock Identifying framing bias in online news.
\newblock \emph{ACM Transactions on Social Computing}, 1(2):1--18.

\bibitem[{Mostafazadeh et~al.(2016)Mostafazadeh, Grealish, Chambers, Allen, and
  Vanderwende}]{mostafazadeh2016caters}
Nasrin Mostafazadeh, Alyson Grealish, Nathanael Chambers, James Allen, and Lucy
  Vanderwende. 2016.
\newblock Caters: Causal and temporal relation scheme for semantic annotation
  of event structures.
\newblock In \emph{Proceedings of the Fourth Workshop on Events}, pages 51--61.

\bibitem[{Nguyen et~al.(2015)Nguyen, Tannier, Ferret, and
  Besan{\c{c}}on}]{nguyen-etal-2015-generative}
Kiem-Hieu Nguyen, Xavier Tannier, Olivier Ferret, and Romaric Besan{\c{c}}on.
  2015.
\newblock \href {https://doi.org/10.3115/v1/P15-1019} {Generative event schema
  induction with entity disambiguation}.
\newblock In \emph{Proceedings of the 53rd Annual Meeting of the Association
  for Computational Linguistics and the 7th International Joint Conference on
  Natural Language Processing (Volume 1: Long Papers)}, pages 188--197,
  Beijing, China. Association for Computational Linguistics.

\bibitem[{Ning et~al.(2018)Ning, Wu, Peng, and Roth}]{ning-etal-2018-improving}
Qiang Ning, Hao Wu, Haoruo Peng, and Dan Roth. 2018.
\newblock \href {https://doi.org/10.18653/v1/N18-1077} {Improving temporal
  relation extraction with a globally acquired statistical resource}.
\newblock In \emph{Proceedings of the 2018 Conference of the North {A}merican
  Chapter of the Association for Computational Linguistics: Human Language
  Technologies, Volume 1 (Long Papers)}, pages 841--851, New Orleans,
  Louisiana. Association for Computational Linguistics.

\bibitem[{Pichotta and Mooney(2016)}]{pichotta}
Karl Pichotta and Raymond Mooney. 2016.
\newblock \href
  {https://www.aaai.org/ocs/index.php/AAAI/AAAI16/paper/view/12157} {Learning
  statistical scripts with lstm recurrent neural networks}.

\bibitem[{Recasens et~al.(2013)Recasens, Danescu-Niculescu-Mizil, and
  Jurafsky}]{recasens2013linguistic}
Marta Recasens, Cristian Danescu-Niculescu-Mizil, and Dan Jurafsky. 2013.
\newblock Linguistic models for analyzing and detecting biased language.
\newblock In \emph{Proceedings of the 51st Annual Meeting of the Association
  for Computational Linguistics (Volume 1: Long Papers)}, pages 1650--1659.

\bibitem[{Rudinger et~al.(2015)Rudinger, Rastogi, Ferraro, and
  Van~Durme}]{rudinger-etal-2015-script}
Rachel Rudinger, Pushpendre Rastogi, Francis Ferraro, and Benjamin Van~Durme.
  2015.
\newblock \href {https://doi.org/10.18653/v1/D15-1195} {Script induction as
  language modeling}.
\newblock In \emph{Proceedings of the 2015 Conference on Empirical Methods in
  Natural Language Processing}, pages 1681--1686, Lisbon, Portugal. Association
  for Computational Linguistics.

\bibitem[{Schank and Abelson(1977)}]{schank1977scripts}
Roger~C Schank and Robert~P Abelson. 1977.
\newblock \emph{Scripts, plans, goals, and understanding: An inquiry into human
  knowledge structures}.
\newblock Psychology Press.

\bibitem[{See et~al.(2017)See, Liu, and Manning}]{CNNDM}
Abigail See, Peter~J. Liu, and Christopher~D. Manning. 2017.
\newblock \href {https://doi.org/10.48550/ARXIV.1704.04368} {Get to the point:
  Summarization with pointer-generator networks}.

\bibitem[{Shen et~al.(2021)Shen, Zhang, Ji, and Han}]{shen-etal-2021-corpus}
Jiaming Shen, Yunyi Zhang, Heng Ji, and Jiawei Han. 2021.
\newblock \href {https://doi.org/10.18653/v1/2021.emnlp-main.441} {Corpus-based
  open-domain event type induction}.
\newblock In \emph{Proceedings of the 2021 Conference on Empirical Methods in
  Natural Language Processing}, pages 5427--5440, Online and Punta Cana,
  Dominican Republic. Association for Computational Linguistics.

\bibitem[{Wang et~al.(2022)Wang, Song, Zhang, Jin, Cho, Yao, Wang, Chen, and
  Yu}]{feiwang}
Fei Wang, Kaiqiang Song, Hongming Zhang, Lifeng Jin, Sangwoo Cho, Wenlin Yao,
  Xiaoyang Wang, Muhao Chen, and Dong Yu. 2022.
\newblock \href {https://doi.org/10.48550/ARXIV.2210.12330} {Salience
  allocation as guidance for abstractive summarization}.

\bibitem[{Wang et~al.(2020)Wang, Chen, Zhang, and Roth}]{wang2020joint}
Haoyu Wang, Muhao Chen, Hongming Zhang, and Dan Roth. 2020.
\newblock Joint constrained learning for event-event relation extraction.
\newblock \emph{arXiv preprint arXiv:2010.06727}.

\bibitem[{Wanzare et~al.(2016)Wanzare, Zarcone, Thater, and
  Pinkal}]{wanzare-etal-2016-crowdsourced}
Lilian D.~A. Wanzare, Alessandra Zarcone, Stefan Thater, and Manfred Pinkal.
  2016.
\newblock \href {https://aclanthology.org/L16-1556} {A crowdsourced database of
  event sequence descriptions for the acquisition of high-quality script
  knowledge}.
\newblock In \emph{Proceedings of the Tenth International Conference on
  Language Resources and Evaluation ({LREC}'16)}, pages 3494--3501,
  Portoro{\v{z}}, Slovenia. European Language Resources Association (ELRA).

\bibitem[{Weber et~al.(2020)Weber, Rudinger, and
  Van~Durme}]{weber-etal-2020-causal}
Noah Weber, Rachel Rudinger, and Benjamin Van~Durme. 2020.
\newblock \href {https://doi.org/10.18653/v1/2020.emnlp-main.612} {Causal
  inference of script knowledge}.
\newblock In \emph{Proceedings of the 2020 Conference on Empirical Methods in
  Natural Language Processing (EMNLP)}, pages 7583--7596, Online. Association
  for Computational Linguistics.

\bibitem[{Yano et~al.(2010)Yano, Resnik, and Smith}]{yano2010shedding}
Tae Yano, Philip Resnik, and Noah~A Smith. 2010.
\newblock Shedding (a thousand points of) light on biased language.
\newblock In \emph{Proceedings of the NAACL HLT 2010 Workshop on Creating
  Speech and Language Data with Amazon’s Mechanical Turk}, pages 152--158.

\bibitem[{Yu et~al.(2020)Yu, Yin, and Roth}]{yu2020paired}
Xiaodong Yu, Wenpeng Yin, and Dan Roth. 2020.
\newblock Paired representation learning for event and entity coreference.
\newblock \emph{arXiv preprint arXiv:2010.12808}.

\bibitem[{Yuan et~al.(2018)Yuan, Ren, He, Zhang, Geng, Huang, Ji, Lin, and
  Han}]{yuan}
Quan Yuan, Xiang Ren, Wenqi He, Chao Zhang, Xinhe Geng, Lifu Huang, Heng Ji,
  Chin-Yew Lin, and Jiawei Han. 2018.
\newblock \href {https://doi.org/10.1145/3269206.3271674} {Open-schema event
  profiling for massive news corpora}.
\newblock In \emph{Proceedings of the 27th ACM International Conference on
  Information and Knowledge Management}, CIKM '18, page 587–596, New York,
  NY, USA. Association for Computing Machinery.

\end{thebibliography}
\bibliographystyle{acl_natbib}

\end{document}